# Semantic Construction Grammar: Bridging the NL / Logic Divide

Dave Schneider, Michael Witbrock
Cycorp, Inc.
7718 Wood Hollow Dr., #250
Austin, TX 78759, USA
+1 (512) 342-4000
daves@cyc.com, witbrock@cyc.com

## ABSTRACT
In this paper, we discuss Semantic Construction Grammar (SCG), a system developed over the past several years to facilitate translation between natural language and logical representations. Crucially, SCG is designed to support a variety of different methods of representation, ranging from those that are fairly close to the NL structure (e.g. so-called 'logical forms'), to those that are quite different from the NL structure, with higher-order and high-arity relations. Semantic constraints and checks on representations are integral to the process of NL understanding with SCG, and are easily carried out due to the SCG's integration with Cyc's Knowledge Base and inference engine [1] , [2].

## Categories and Subject Descriptors
I.2.7 [Artificial Intelligence] Natural Language Processing – *language parsing and understanding*

## General Terms
Algorithms

## 1. INTRODUCTION
The Cyc SCG semantic interpreter aims to use existing knowledge actively in extracting detailed semantic representations from natural language, which can then be used for querying or extending a semantic knowledge base. In short, we do this by describing SCG constructions: semantically constrained structures in a natural language (like English), for which logically composable and inferentially productive logical representations can be produced. The approach, which was inspired both by work in the 1990s by Charles Fillmore on FrameNet [3], and at CMU on Example Based Machine Translation [4], is, in its first implementation, non-syntactic; it relies on finding sequences of semantically understood elements and surface word forms for which a precise semantics can be assigned. In SCG, precise semantics are assigned to text by composing elements represented in the CycL language, which offers a good combination of representational power and vocabulary, and is supported by the Cyc Inference Engine and Knowledge Base, which are both used for semantic checks during parsing. Cyc contains a wealth of real-world concepts and relationships between them [2], and is able to use them to perform complex inference in a variety of domains using a variety of specialized reasoners to supplement general theorem proving.

One key feature of the Semantic Construction Grammar system discussed here is the ability to output logical representations in a wide variety of styles, from those that are very similar to the syntactic form to those that are quite different. Below is a very simple example of a phrase with a straightforward recursive compositional interpretation:

(1) "Big blue building"
    (LargeFn
      (SubcollectionOfWithRelationToFn Building
        mainColorOfObject BlueColor))

Interpretation of this phrase proceeds in a bottom up manner, starting with "blue building", with that interpretation substituted into the logical template from the construction that consumes

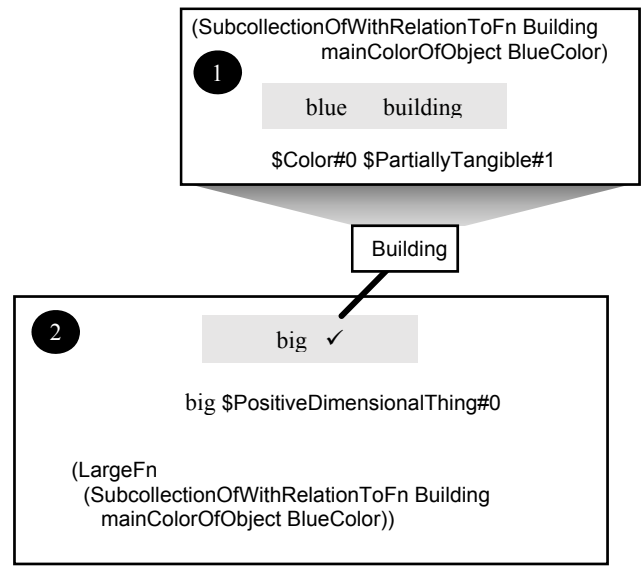

**Figure 1: Interpretation of "big blue building".** Box 1 shows the logical representation that results from matching "blue building" to the NL template "$Color#0 $PartiallyTangible#1". The resulting representation is logically typed as Building, which serves as one of the inputs matching the second construction specifying the word "big" followed by a PositiveDimensionalThing, of which Building is an indirect specialization.

"big". A schematic version of this recursive interpretation is shown in Figure 1.

In addition to relatively straightforward examples like (1) above, the relationship between the NL arguments can be quite complex as in the resulting logical representation:

(2) "Submarine base"
    (SubcollectionOfWithRelationToFn **MilitaryBase-Grounds**
      (Kappa (?VAR1 ?VAR2)
        (behaviorCapable ?VAR1 ?VAR2 fromLocation))
      (DeployingMaterialOfTypeFn **Submarine**)))[1]

The examples above illustrate another key feature of the SCG system—it is not restricted to specific semantic domains or sets of relationships. Instead, it is able to operate on any domain for which appropriate constructions have been specified, and can use constructions from multiple domains simultaneously.

Other systems that attempt to understand similarly wide-ranging domains include Johan Boxer [5], and derived tools (e.g. [6]).

## 2. CONSTRUCTIONS

A hallmark of the SCG system is the ability to treat both common compositional patterns and idiomatic expressions in a uniform manner. In either case, the NL template of a construction matches the input, and the resulting semantics are produced. In essence, everything is treated as an idiomatic construction, though some are more general than others.

### 2.1 Natural Language Templates

The NL template of an SCG construction consists of a sequence of strings and typed variables; it may be a mixture of both, or may consist entirely of one or the other. Sequences may be required, optional, or may contain alternatives. The typed variables (shown as a Cyc term prefixed with $ and suffixed with # and a uniquifying integer[2]) are variables that must be matched by a sub-interpretation with either the same type as the term in the variable, or a more specific type. A few examples are shown below:

(3) $Color#0 $PartiallyTangible#1
(4) big $PositiveDimensionalThing#0
(5) the [last|most recent|latest] $Event#0 was in [$Date#1 in $Place#2|in $Place#2 in $Date#1]
(6) kick the bucket

In (5), one of "last", "most recent", or "latest" is required, as is one or the other of "$Date#1 in $Place#2" or "in $Place#2 in $Date#1". (6) is an example of an NL template that has no variables in it. Such templates are suitable for some idiomatic expressions, and can also be used as a substitute for making an entry in the lexicon used for concept tagging (see discussion below in 3.1)

---

[1] This expression denotes the collection of military bases that are capable of filling the fromLocation role in a submarine deployment. Kappa is used to reduce the arity of behaviorCapable from ternary to binary by fixing the fromLocation argument. The Kappa term denotes a binary predicate where the first argument (the base) is capable of being the fromLocation in an instance of the second (submarine deployment).

[2] The uniquifying number keeps identically-typed terms unique within a construction. For consistency, the number is always required, even when there is no ambiguity in a particular construction.

Each NL template is associated with a language. A single construction can contain multiple NL templates, which may be in multiple languages. For example, one construction related to cooking contains both an English template and a French template:

(7) place[d|{}] $ContainerArtifact#0 over high heat
(8) placer $ContainerArtifact#0 à feu vif

The English template (7) also shows how morphological variants of fixed elements are currently handled: explicit enumeration.

In order to ease the search burden when finding constructions, the cross product of the individual variants are stored. In the case of (5) above, six different variants would be stored. Additionally, simplified variants of each construction are also stored, which allow for dramatic pruning of the space that needs to be searched when retrieving constructions (cf. 3.2 below).

---

NL Template: $AminoAcid#0$PositiveInteger#1$AminoAcid#2-$PolypeptideMolecule#3

Skeleton Template: $Thing$Thing$Thing-$Thing

Lexical Template: -

---

**Figure 2:** A sample NL template (the one used to interpret the string "G12V-K-Ras" from molecular biology). Typed variables are effectively collapsed into untyped variables in the skeleton templates, while all variables are completely removed from the lexical templates, leaving only the fixed strings (only "-" here).

### 2.2 Logic Templates

SCG constructions may contain multiple NL templates, but each contains exactly one logic template. This corresponds to the idea that there may be multiple realizations of the same semantic content, but that logic should be, by its very nature, unambiguous. The logic template is simply a CycL logical expression, which may be a logical sentence or may denote something non-sentential (e.g. the concept "Color"). When a construction is used by the SCG interpreter, the resolved interpretations of each of the typed variables is substituted into the logic template in place of the variable.

(9) (LargeFn $PositiveDimensionalThing#0)

Example (10) shows just how complicated the relationship can be between the typed variables.

(10) (CollectionSubsetFn $Food#1
        (TheSetOf ?FOOD
          (and
            (isa ?ART $Food#1)
            (intendedSoleFunction ?ART
              (SubcollectionOfWithRelationToTypeFn EatingEvent doneBy $Animal#0) consumedObject))))

This is the logic template used to relate an animal with a type of food intended solely for that animal (e.g. "cat kibble").

#### 2.2.1 Anaphoric References

In most constructions, the typed variables found in the logic template are mirrors of those in the NL template. However, constructions also contain a field for anaphoric references that need to be resolved from elsewhere in the context (e.g. from elsewhere in the NL expression or from background knowledge). For example, (11) shows the construction that might be used for "the end of the 2015 season" in an article about sports.

(11) "[the|{}] end of the $Integer#0 season"
      (#$EndFn
        (#$AnnualEventOfYearFn
          (#$SeasonOfSportEventTypeFn $SportsEvent#1)
          (#$YearFn $Integer#0))))

This CycL representation includes reference to the specific sport, but there's no mention of a sport in the NL template. When the construction is used by the SCG interpreter, the presence of $SportsEvent#1 in the anaphoric references field triggers a contextual search for an appropriate reference along compatible parse paths.

*2.2.2 Output Type and Variable*

In some cases, the resolved logic template should simply be substituted into any "matrix" template which satisfies the relevant type constraints. Occasionally, though, what should be substituted is a variable from the logic pattern, which is then conjoined with the matrix. Which variable should be used for this assembly is specified in the construction. Similarly, because the logical expression, as completed, may have a very complex type (generalizing to many types, or having a type that is functionally described), it is possible to specify a type that should be used for matching for insertion. This may either be an explicit type, or one of the typed variables like $SportsEvent#1 in which case the interpreted type from the NL expression is used.

## 2.3 Semantic Tests

In addition to the constraints imposed by the typed variables in the NL template, arbitrarily complicated logical tests can be added to a construction. There are two types of tests: positive and negative. Positive tests add constraints, such as additional type constraints, or a relationship between two arguments that must hold. Consider the following construction:

(12) "$ActionOnObject#0 $ExistingObjectType#1"
  (SitTypeSpecWithTypeRestrictionOnRolePlayerFn
    $ActionOnObject#0
    objectActedOn
    $ExistingObjectType#1)

This construction can be applied to phrases like "blowing out candles", and would mean something like "the blowing-out events in which candles are acted on". This construction contains an additional positive semantic test specifying that the objectActedOn must be capable of being acted on in this type of event:

(13) ((TypeCapableFn behaviorCapable)
    $ActionOnObject#0
    objectActedOn
    $ExistingObjectType#1)

Given that "blow out" can refer to several different types of events (cf. "I blew out a tire", "the explosion blew out a window", "she blew out the candles"), this semantic test ensures that only the appropriate event types are used for the different sorts of entities.

Negative constraints prevent the construction from being applied, even when the type constraints in the NL template are satisfied. Typically, negative constraints carve out a small segment of the ontological space for which a construction does not apply.

(14) "$PartiallyTangible#0 $MovementEvent#1"
  (SitTypeSpecWithTypeRestrictionOnRolePlayerFn
    $MovementEvent#1
    toLocation
    $PartiallyTangible#0)

This construction can be used to interpret "intracellular accumulation" to mean movement in the interior of cells. However, it can also be used to interpret "electron transport" as transportation into electrons. While this is, arguably, logically possible, it is not a standard interpretation of the phrase, and is probably physically impossible. A negative constraint that restricts the type of the PartiallyTangible to not be an instance of sub-atomic particle can be used to prevent that reading[3]:

(15) ¬(genls $PartiallyTangible#0 SubAtomicParticle)

## 3. SEMANTIC INTERPRETATION

The semantic interpretation process used by the SCG involves: (1) conceptually tagging the text input with relevant Cyc concepts, (2) retrieving possible constructions, (3) applying the constructions and semantically testing their output, and (4) composing the results of recursive construction application.

## 3.1 Concept Tagging

The first step in semantic interpretation involves performing concept tagging on the textual input. The text is sent to the Cyc concept tagger, which analyzes the input and adds concept tags for each phrase that it's able to recognize. The concept tags may be for overlapping and/or subsuming regions. For ambiguous words, multiple concept tags may be added to a single text span. If multiple concepts for the same span are used in interpretations, multiple interpretations of that span are carried forward (i.e. there's no attempt at consolidation). Additionally, in some cases, portions of words will receive their own concept tags. For example, in the case of "G12V-K-Ras" (a kind of mutated protein), the following concepts are tagged as possible meanings:

**Table 1: Available Concept Tags for "G12-V-K-Ras"**

| G | Glycine |
| | gibbsFreeEnergyOfSystem |
| | Gram |
| | GuanineDeoxyribonucleotide |
| | (AminoAcidResidueTypeFn Glycine) |
| | GeneralRating |
| 12 | 12 |
| V | Volt |
| | Valine |
| | V-TheTVMiniSeries |
| | (AminoAcidResidueTypeFn Valine) |
| K-Ras | K-Ras-Protein |

Because of the many semantic constraints that are built into the interpretation system (e.g. typed variables, semantic tests, etc.), the presence of so many concepts doesn't generally cause serious problems, and the SCG is able to interpret the word "G12V-K-Ras" uniquely as:

(16) (PolypeptideTypeWithResidueAtPositionReplacedByResidueTypeFn
    K-Ras-Protein
    (AminoAcidResidueTypeFn Glycine) 12
    (AminoAcidResidueTypeFn Valine))

## 3.2 Construction Retrieval

In order to determine which constructions should be applied, the system runs a window over the input text. The maximum length of that window is variable, but is typically set at about 10-12 tokens, where tokens include any word (or sub-word that receives a concept tag). If a construction can be applied to the entire window, it is applied and the window moves one token to the right. If no constructions can be applied, the window is shrunk by one token and constructions are searched for anew. This process continues until either one or more constructions can be applied or the window shrinks to zero tokens.

---

[3] Many additional negative constraints could be used to restrict the construction from applying to other phrases. Since there's no restriction on the number of positive or negative constraints, they could be easily added; there is an obvious opportunity for machine learning in acquiring these constraints. We plan to pursue constraint induction in future work.

The process of actually retrieving applicable constructions for a particular window involves checking the cross-product of each of the available interpretations for each token (as well as the token itself, in case it's mentioned as a string in an NL template). In the case of (16), that would involve 140 (7*2*5*2) possible patterns. However, because the constructions admit any term that is known to be a specialization of their typed variables, the system in fact needs to search for the cross product of the generalizations of each of the terms, not just the cross product of the tagged concepts. Because the Cyc KB is rich and detailed, most nodes in the ontology have 40-120 generalizations ('generalization' here refers to both isa and genls ontology links in the Cyc ontology, corresponding to instance typing and type generalization respectively). Assuming that each concept has only 40 generalizations, approximately 6,144,000 ((6*40)*(1*40)*(4*40)*(1*40)) patterns would have to be searched for just this particular window. Even with a modern, indexed database, searching for so many patterns causes a substantial slowdown.

In order to prune the search space for each window, a number of optimizations have been applied. Instead of immediately searching all 6.1 million possible expansions for the phrase "G12V-K-Ras", the SCG first produces all possible purely lexical templates for the window (n=$i^2$/2 possibilities, where $i$ is the number of tokens, which is typically limited to 10-12) and checks the template repository to see if there are constructions that match. For those lexical templates that have matches, the system produces a set of possible skeletons, and checks to see which of those match some construction. Finally, for each skeleton that matches, the cross-product of the generalizations of the concepts in the text is produced, using only generalizations present as typed variables in some template, and the construction repository is searched for constructions that exactly match. All matching constructions are retrieved and applied to generate candidate logical expressions. All possible such interpretations for each text span are produced and carried forward to semantic testing.

### 3.3 Semantic Tests

Once a candidate construction has been identified for a window, its positive and negative semantic tests (discussed in 2.3 above) are checked. If any of these semantic checks fails, the construction is discarded.

Once the construction-specific semantic tests have passed, the proposed interpretation is subject to a further set of plausibility constraint checks. These checks focus on testing whether or not the representation meets all the constraints imposed by the Cyc KB. One of the most common problems encountered is a type/instance mismatch. In the Cyc KB, there is a strong distinction between instances of a collection and specializations of a collection. For example, Skyscraper is a specialization of Building, while TheWhiteHouse is an instance. Plausibility checking is used the throw out the interpretation of "White House dancing" as "dancing that's to the White House" that would be produced by construction (14) above. TheWhiteHouse meets the basic constraint of PartiallyTangible, but the relations in that construction have a KB requirement that the PartiallyTangible argument be a specialization of PartiallyTangible, not, like the White House, an instance of it.

Additional features of plausibility testing include the ability to remove interpretations that are known to be false, and those that violate argument restrictions, including inter-argument restrictions. For example "a bank is a kind of company" cannot be interpreted as (genls Bank-Topographical Business), because riverbanks are known to be disjoint with companies.

Applicability of inter-argument restrictions can be seen through the following:

(17) The song has 6 notes.

It's clear that "notes" here is intended to refer to MusicalNote, not Note-Document. However, if the predicate properPartTypeCount is used for the representation, its argument restrictions are insufficient to prevent this reading, since intangible objects can have intangible parts, just as tangible object can have tangible parts. The reading of Note-Document could be ruled out during plausibility checking by appeal to the following inter-argument restriction:

(18) (interArgGenl1-2 properPartTypeCount Intangible Intangible)
*If the first argument to properPartTypeCount is wholly intangible, the second must also be wholly intangible.*

All of the semantic tests are performed in a specific application context that inherits the general constraints in the Cyc KB; application or domain-specific constraints can be added to a specific application context when appropriate. Every interpretation that passes the semantic tests is admitted for further processing.

### 3.4 Interpretation Composition

As mentioned in 2.2.2, each lexical entry and construction has an output variable which is used when composing it into subsuming constructions. The basic process for composing interpretations is to substitute the output variable of the sub-interpretation into the logical template of the super-construction, and then to conjoin the two interpretations together. Interpretations that wouldn't otherwise have variables are assigned to a variable[4]. A straightforward logical simplification happens at the end, wherein terms are substituted in for variables they are equal to, and redundant and connectives are simplified away. The entire interpretation composition process is shown in Figure 3.

Once a construction has been successfully applied and the arguments have all been successfully composed, a new edge is added to the parse graph, and that edge triggers a new search for possible constructions using it.

### 3.5 Anaphor Resolution

When the system uses a construction that contains an anaphoric reference, it attempts to resolve the referent. As mentioned earlier, the resolution system is not inherently limited to finding linguistic referents. The current implementation searches back through the parse graph, looking for any phrase with an interpretation that meets the typed variable constraint. The search process stops when five possible antecedents have been found or the beginning of the graph has been reached, whichever comes first.

While this implementation is extremely simplistic, it performs surprisingly well—semantic constraints provide enough focus to substantially constrain the possible antecedents. However, we

---

[4] For expository purposes, these have been left off lexical entries and logical representations elsewhere in this paper.

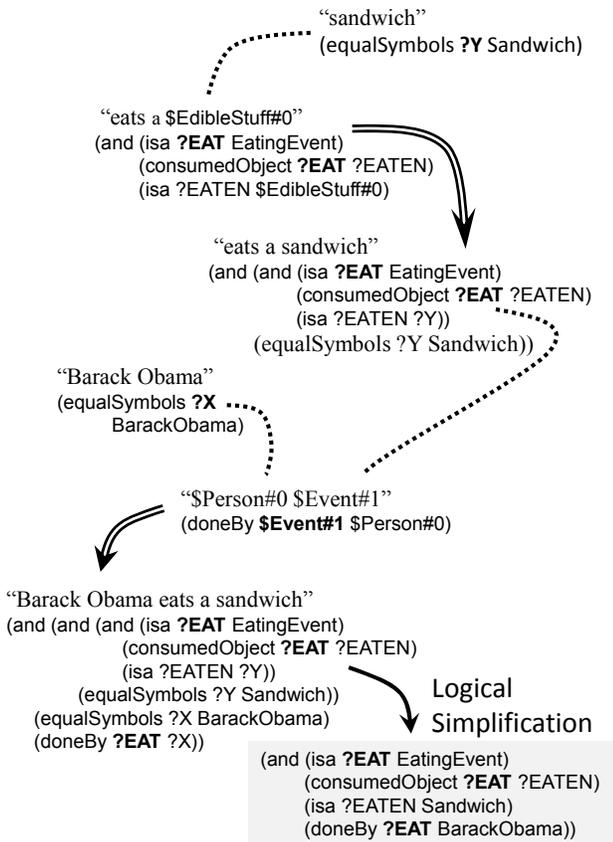

fully acknowledge that this is far from a fully-fledged anaphor resolution system. External anaphor resolution systems could be easily integrated into the SCG system to provide more constraints on which phrases are possible antecedents. In the current system, the exception checking mechanism can be used prevent anaphoric

**Figure 3: Example Interpretation Composition for "Barack Obama eats a sandwich".** Dotted lines indicate terms that are substituted into another logical template. Double line arrows link a construction being substituted into to the resulting logical representation. The final step shows the logical simplification that takes place at the end of each chain of interpretation. Bold variables indicate the output type at each step.

matches to an unlikely but compatible type.

## 4. CHALLENGES

The use of detailed, precise logical representations as a semantic target presents a number of challenges, some of which are detailed below.

### 4.1 Coordination

Coordination is rampant in natural language text, but presents significant challenges to semantic interpretation systems. The challenges are magnified for systems that attempt to produce precise semantics over heterogeneous representations, since a single way of dealing with the semantics of coordination is unlikely to work for all cases. The problem of coordination is also made more difficult by the fact that the logical representations used by this system don't always line up nicely with the linguistic constituent structures that are typically coordinated. For example, if a construction uses the subject, verb, and direct object of a sentence, it will be difficult to produce coordinate semantics for a sentence of the form "S V and V DO", such as "the man bought and ate a sandwich". Even when the constructions align with what needs to be coordinated, production of the combined logical representation is not completely straightforward. The most naïve representation of "bought and ate a sandwich" might look like this:

(19) (and (isa ?EVT
          (CompositeActivityFn (TheSet EatingEvent Buying)))
     (objectActedOn ?EVT ?SAND)
     (isa ?SAND Sandwich))

This is not, however, a particularly good representation for the coordination in "bought and ate a sandwich" because of the very weak role that the sandwich plays in the composite event. A better representation would have the sandwich playing different, more specific roles in the two events:

(20) (and (isa ?EAT EatingEvent)
         (isa ?BUY Buying)
         (objectPaidFor ?BUY ?SAND)
         (consumedObject ?EAT ?SAND)
         (isa ?SAND Sandwich))

*4.1.1 Collective vs. Distributive Interpretation*

While precise logical representations present difficulties for some aspects of coordination, they will also provide resources for solving other coordination issues, in particular the question of whether collective or distributive readings are even possible. For example, in "Fred and George answered the phone", there are two different phone answering events. The knowledge that PickingUpATelephoneReceiver is a SingleDoerAction can allow the system to automatically discard the reading in which there was a single event in which Fred and George both picked up the same telephone. The same sort of knowledge about how many doers there are in events would allow the system to interpret "Fred and George woke" as involving two separate WakingUpFromSleep events, not just one that they both carried out.

### 4.2 Quantification

The precise logical representations produced by SCG make it easy to express many types of quantification (e.g. universal, existential, numeric, etc.). This does not, however, lessen the difficulty of the task of determining which quantification should be used in a logical translation. While SCG does not currently have a full treatment of quantification, certain types of quantification are handled. In particular, numeric quantifiers on entities are generally interpreted with constructions that compose a number (including imprecise numbers like "a few") with an entity type, resulting is a group of *n* entities. That group is then used as the interpretation of the numerically quantified phrase.

(21) "2 sandwiches"
    (SubcollectionOfWithRelationToFn
     (GroupFn Sandwich) groupCardinality 2))
    *subcollection of groups of sandwiches, each of which has a cardinality of two*

Even after numerically quantified entities have been understood, they present additional difficulties, such as the need to determine if they should be interpreted with a collective or distributive reading.

The treatment of variables that remain unquantified at the outermost interpretation is treated as an application-dependent feature. For statements, variables are typically existentially quantified before the resulting sentence is asserted into the Cyc KB. For questions, the variables can remain unquantified, which results in a query to determine the actual bindings for the variables. In other applications, it is sufficient to merely determine whether any such bindings exists, in which case the unquantified variables are existentially bound.

## 4.3 Evaluation

Performing a fair evaluation of a semantic interpretation system like this one poses substantial challenges. The biggest of these is that the space of representations is very large, and the actual representations produced are quite detailed and specific. This makes production of a gold standard quite difficult, since the same content can very reasonably be represented in a number of different ways, all of which are equally valid, but which cannot be trivially transformed into a single canonical representation. This means that the common practice of providing a single gold-standard representation will penalize systems that allow for this sort of representational flexibility. The simplest sort of evaluation might involve determining the percentage of text that is covered by some sort of compositional interpretation. While this sort of evaluation is tempting, its value is at best questionable. In internal work, that number has proved very easy to skew with just a few overly productive constructions. The result is that the percentage of text understood can very easily inversely correlate with precision.

Most existing evaluations of semantic annotations (e.g. SemEval 2013 spatial role labeling [7], BioNLP 2013 [8]) are effectively testing whether a particular relationship holds between two or more elements of the text, and those evaluation systems are designed to determine whether or not the correct relation is found between the exact text segments identified in the gold-standard data. Some evaluations (e.g. BioNLP 2013) have simplified this problem somewhat by providing pre-annotated entities as part of the training and test data, so the annotator can concentrate on finding the right relationships and events. While pre-identification of the relevant entities makes the evaluation task simpler (in that it can be more easily automated), it is an artificial constraint that is unrealistic in real-world applications, where there is no agreed-upon set of entities. Similarly, identifying the source text spans for semantic objects is of no intrinsic benefit to understanding, and presents a substantial difficulty, since the form of the logical (resulting) representation may not correspond in any direct way to portions of the text string.

While we have not yet carried out a rigorous evaluation of the SCG system, we are in the process of developing one. This evaluation will use a portion of the Flickr 8K data set [9], which is a set of 8000 images from Flickr (originally collected for machine vision evaluations) that were manually annotated with five captions each. A subset of these captions will be run through the SCG interpreter, and the results will be manually scored. The scoring will be carried out starting with the interpretations covering the largest segment of the input. Each of those interpretations will be manually checked for correctness (i.e. does it correctly relate the entities mentioned in the interpreted text). If no correct interpretations are found, the next largest string with an interpretation will be evaluated. The process will continue until there is a correct interpretation or until there are no more compositional interpretations for the caption.

Metrics to be calculated over this data will include at least the following: percentage of input text with a correct interpretation, percentage of interpretations that are correct (precision), and mean length of correct interpretations.

## 5. CONCLUSION

Semantic Construction Grammar represents an aggressive attempt to produce precise, detailed logical representations that is not limited to narrow domains. The use of such a logical representation presents opportunities for innovative methods of semantic disambiguation and pruning that are not available to systems using shallow semantic representations or which are not backed up by a large knowledge base and capable inference engine.

At the same time, the space of possible logical representations presents additional challenges that are not generally present for systems using weaker or less precise semantics. These challenges include the need to thoroughly understand quantification and coordination, and the presence of potentially very many possible interpretations which must be sorted through, and hopefully eliminated. Additionally, evaluation of such a system is substantially challenging, due to the many possible representations. Not only are multiple equivalent representations possible, but the amount of information that could be extracted is very large; producing even a single representation of modest amounts of text is a daunting task.

We look forward to addressing these challenges and expanding the scope of the Semantic Construction Grammar.